\documentclass{article}
\usepackage{spconf,amsmath,epsfig}
\usepackage{graphicx}
\usepackage{multirow}
\usepackage{times}
\usepackage{amssymb}
\usepackage{amsfonts}
\usepackage{color}
\usepackage{subfigure}
\usepackage{epstopdf}
\usepackage{enumitem}

\usepackage{adjustbox,lipsum}
\usepackage{dsfont}
\usepackage{booktabs}
\usepackage{verbatim}
\usepackage[font=small,skip=0pt]{caption}
\begin{document}\sloppy

% Example definitions.
% --------------------
\def\x{{\mathbf x}}
\def\L{{\cal L}}
% Title.
% ------

\title{A Mask Based Deep Ranking Neural Network for Person Retrieval}

% \author[1]{Lei Qi}
% \author[1]{Jing Huo}
% \author[2]{Lei Wang}
% \author[1]{Yinghuan Shi}
% \author[1]{Yang Gao}
% \affil[1]{State Key Laboratory for Novel Software Technology, Nanjing University}
% \affil[2]{School of Computing and Information Technology, University of Wollongong}

\name{Lei Qi$^{1}$, Jing Huo$^{1}$, Lei Wang$^{2}$, Yinghuan Shi$^{1}$, Yang Gao$^{1}$\thanks{This work was supported by the National Science Foundation of China (Grant Nos. 61432008, 61673203, 61806092), Jiangsu Natural Science Foundation (BK20180326), the Young Elite Scientists Sponsorship Program by CAST (YESS 2016QNRC001), CCF-Tencent Open Research Fund and the Collaborative Innovation Center of Novel Software Technology and Industrialization.}}
\address{$^{1}$State Key Laboratory for Novel Software Technology, Nanjing University\\
$^{2}$School of Computing and Information Technology, University of Wollongong}

% \name{George P. Burdell and John Q. Professor
%                       \thanks{This work was supported by...}}
%                 \address{Common address, department \\
%                          City, etc \\
%                          optional e-mail address}

\maketitle

\begin{abstract}
 Person retrieval faces many challenges including cluttered background, appearance variations (e.g., illumination, pose, occlusion) among different camera views and the similarity among different person's images. To address these issues, we put forward a novel mask based deep ranking neural network with a skipped fusing layer. Firstly, to alleviate the problem of cluttered background, masked images with only the foreground regions are incorporated as input in the proposed neural network. Secondly, to reduce the impact of the appearance variations, the multi-layer fusion scheme is developed to obtain more discriminative fine-grained information. Lastly, considering person retrieval is a special image retrieval task, we propose a novel ranking loss to optimize the whole network. The proposed ranking loss can further mitigate the interference problem of similar negative samples when producing ranking results. The extensive experiments validate the superiority of the proposed method compared with the state-of-the-art methods on many benchmark datasets.
\end{abstract}

\section{Introduction}
\begin{figure}%[!h]
\centering
\includegraphics[width=9cm,height=4cm]{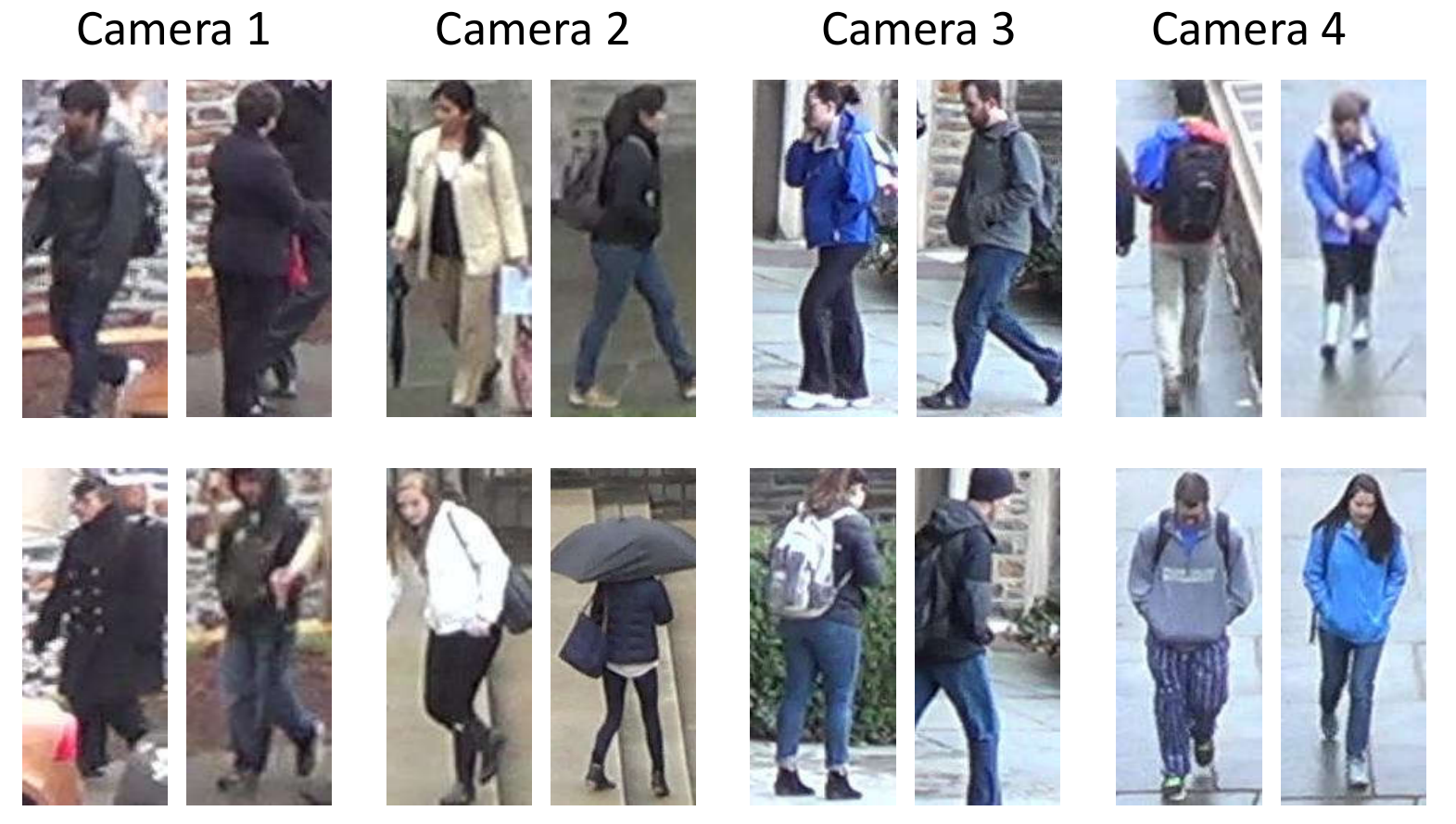} %
\caption{Images from different cameras on DukeMTMC-reID.}
\label{fig5}
\vspace*{-20pt}% µ÷Õû¼ä¾à
\end{figure}

Person retrieval, also known as person re-identification (Re-ID), is to match images of the same individual captured by non-overlapping camera views. There are many challenges in person Re-ID, including cluttered background, appearance variations (e.g., illumination, pose, occlusion, resolution) among different camera views and interference of similar images with different identities. Fig.~\ref{fig5} shows images from different camera views on DukeMTMC-reID~\cite{ristani2016performance}. As seen, the images in the same camera view have similar background, while the background differs in different camera views. In addition, appearance variations, such as illumination and resolution, also lead to noise in the extracted person feature representation. Moreover, since person Re-ID can be seen as an image retrieval task, there is an interference problem of similar negative samples in the retrieval precess.

%\vspace*{-10pt}%,

\begin{figure*}%[!h]
\centering
\includegraphics[width=15cm,height=3.5cm]{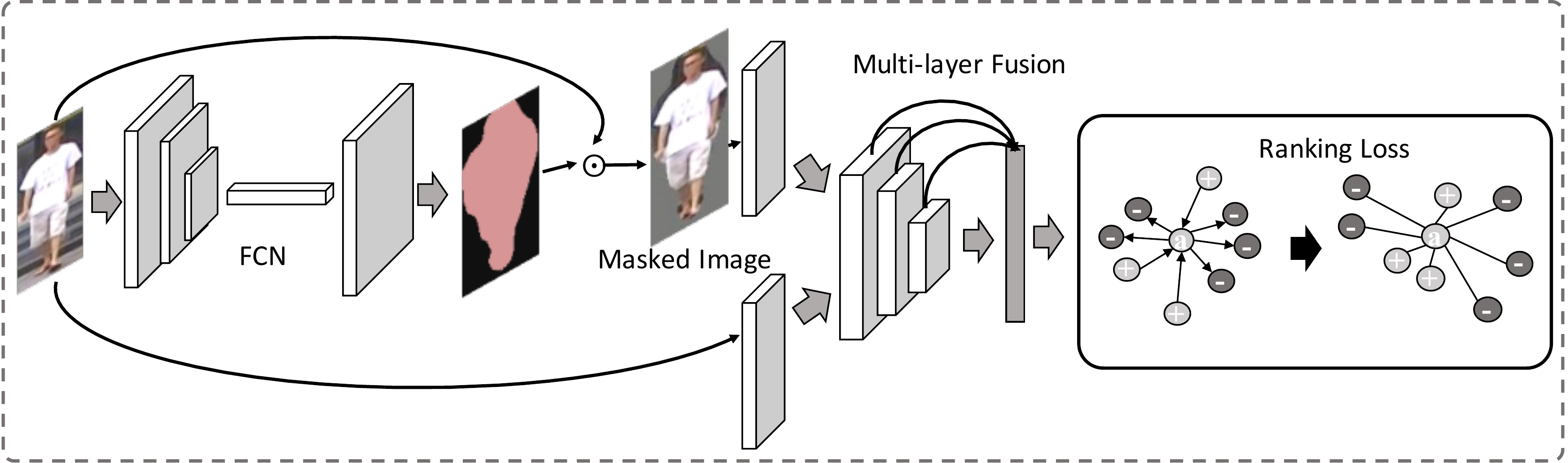} %
\caption{The pipeline of the proposed method for person retrieval. The image segmentation network is firstly employed to obtain the segmentation results of a person image. Then a mask based deep ranking neural network with a skipped feature fusion layer is proposed to extract robust person features. Finally, a novel ranking loss is designed to train the overall deep neural network.}
\label{fig3}
\vspace*{-15pt}% µ÷Õû¼ä¾à
\end{figure*}

Many efforts have been made to address the challenges in person retrieval~\cite{DBLP:conf/eccv/VariorSLXW16}\cite{DBLP:conf/cvpr/XiaoLOW16}\cite{DBLP:conf/cvpr/ZhaoTSSYYWT17}\cite{DBLP:conf/cvpr/ChengGZWZ16}\cite{DBLP:conf/cvpr/ZhouWWGZ17}\cite{DBLP:conf/cvpr/ChenCZH17}. However, most existing methods neglect the problem of background clutter which leads to degraded performance. Besides, these methods adopt the general neural network structure for Re-ID~\cite{DBLP:conf/eccv/VariorSLXW16}\cite{DBLP:conf/cvpr/XiaoLOW16}\cite{DBLP:conf/cvpr/ZhaoTSSYYWT17}. However, there are usually large intra-personal appearance variations in person images. It is thus important to design more specific network structures to capture the fine-grained information from person images. Moreover, as person Re-ID is also an image retrieval problem, most existing Re-ID frameworks are optimized by contrastive loss or triplet loss~\cite{DBLP:conf/cvpr/ChengGZWZ16}\cite{DBLP:conf/eccv/VariorHW16}, which employ only one negative sample and one positive sample for an anchor at each iteration. As the ranking process in fact involves a set of samples, using the above loss functions may lead to poor results, which may be interfered by similar negative images.

To address the above mentioned problems, we develop a mask based deep ranking neural network with skipped fusing layer (MaskReID), as shown in Fig.~\ref{fig3}. First, to reduce the impact of cluttered background, an image segmentation network is employed to obtain the segmentation results of a person image. The masked image with removed background is adopted as an additional input for feature extraction. Second, considering that person Re-ID is a special fine-grained image recognition task and the different layers of deep neural networks can extract low-, middle- and high-level features, we employ a skipped feature fusion layer scheme to fuse the multi-layer features in the deep neural network. This strategy can extract invariant person representation from different camera views, as it combines all low-level edge and shape information, middle-level structure information and high-level semantic information together. Besides, it can also better propagate the loss to lower layers of neural network for training, making the learning of low-level and middle-level features more accurate. Last, as person Re-ID is a ranking task, i.e., one person has multiple images from different camera views, a ranking loss function is developed to reduce the impact of similar negative samples when producing ranking results. Particularly, an anchor sample interacts with multiple positive and negative samples via exploiting the proposed ranking loss in each iteration. Together, the above three improvements give rise to a novel deep learning framework for person retrieval. 

In summary, the major contributions in this work include:
i) A novel deep learning framework is proposed, which accepts both the original and masked images as input. Besides, we develop a skipped feature fusion layer to extract robust features;
ii) For the person retrieval task, a novel ranking loss function is proposed to further mitigate the interference issue of similar negative samples;
iii) Extensive experiments on multiple benchmark datasets demonstrate the superiority of the proposed method when compared with the state-of-the-art methods. Moreover, through ablation studies, the efficacy of the proposed masked input, the skipped feature fusion scheme and the ranking loss is also verified.
\section{MASK BASED DEEP RANKING NEURAL NETWORK}
The framework of the proposed mask based deep ranking neural network is shown in Fig.~\ref{fig3}. An image segmentation deep network is employed to capture person (foreground) region and remove the background. Besides, there is a feature fusion layer which merges the multi-level information of the proposed network to extract robust person representation. Lastly, a ranking loss is developed to optimize the proposed network, which aims to rank person images with the same identity at top positions in a training batch.
\subsection{A New Mask Based Deep Neural Network Structure}~\label{sec:framework}
%\vspace*{-25pt}
\textbf{Masked Input.}
Person images, captured under complex scenes (e.g., airport or station), have very messy background. Besides, the images of different identities in the same camera view have similar background, while the background of the same identity differs in different camera views. These factors result in a challenge that the neural network cannot fully focus on foreground regions. The ideal feature extraction module should try to distinguish the silhouette of the person so as to focus more on the person region instead of the background. Thus it can improve the performance of person retrieval. Currently, image segmentation can effectively separate the foreground from the background. In this paper, Fully Convolutional Networks (FCN)~\cite{DBLP:conf/cvpr/LongSD15} is employed to obtain the masked images. Particularly, for low quality images such as low resolution, poor segmentation results could be generated by the segmentation network. To address this issue, we employ both the original and masked images as the input of the proposed network.

\begin{figure}%[!h]
\centering
\includegraphics[width=8cm]{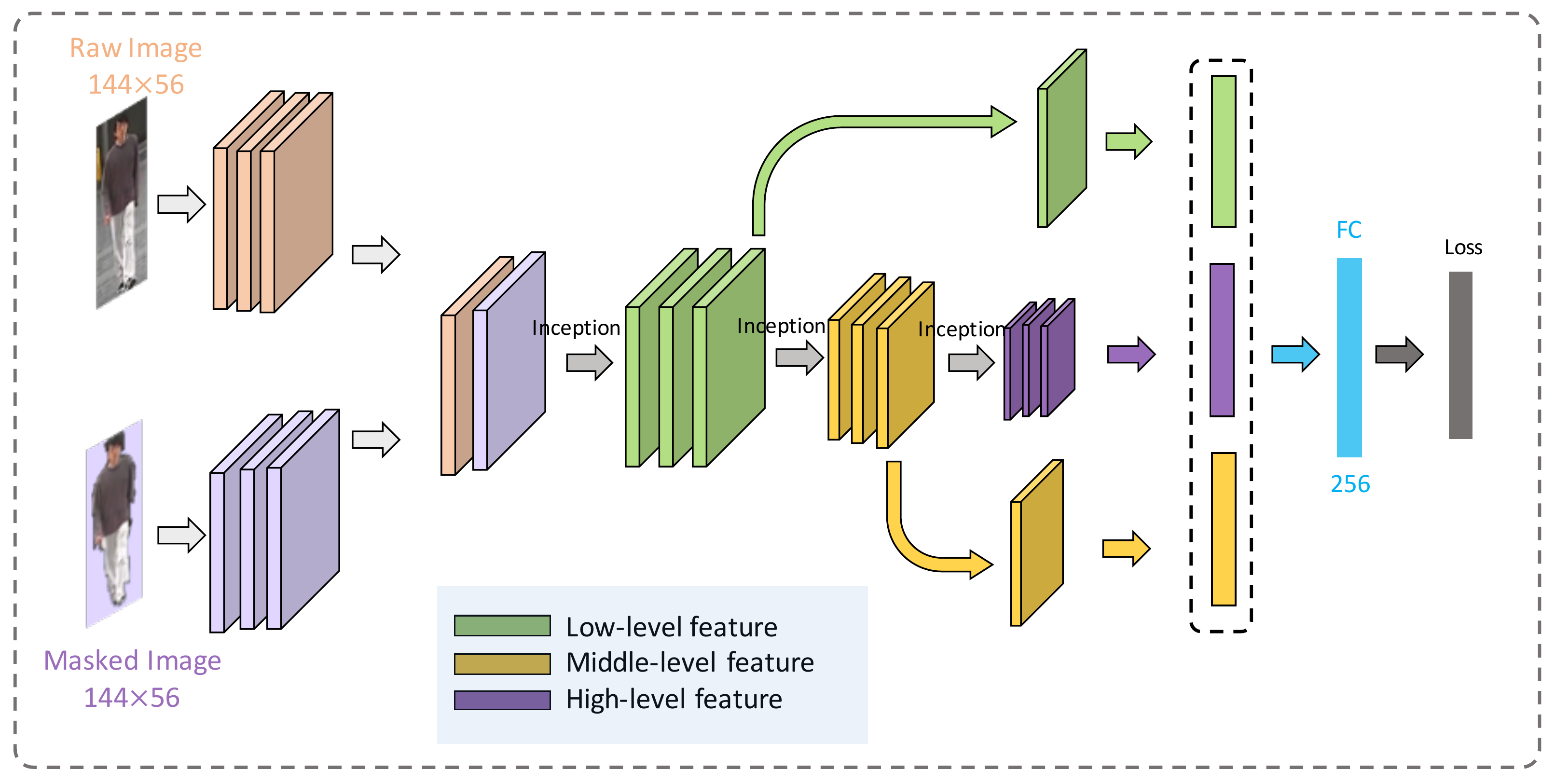}
\caption{Illustration of the proposed framework. First, original and masked images are utilized as input. Second, a multi-layer feature fusion scheme is developed in the proposed network. Last, the network outputs a 256-d (dimensional) feature.}
\label{fig1}
\vspace*{-15pt}% µ÷Õû¼ä¾à
\end{figure}

\textbf{Skipped Feature Fusion Layer Structure.} As for the network structure, the network of Xiao \emph{et al.}~\cite{DBLP:conf/cvpr/XiaoLOW16}, which has an inception structure, is adopted as the basic network. Especially, domain guided dropout, which is designed to activate the neurons from a special domain, is not utilized in the proposed framework. The proposed network structure is shown in Fig.~\ref{fig1}. The two images are processed with several separated layers. Then the two separated features are concatenated and fed into a shared network with three levels, which extract the low-, middle-, and high-level features, respectively. The features of these three levels are fused by a skipped fusing layer to produce the merged feature. The benefits of using different levels of features have been proved in style transfer tasks~\cite{DBLP:conf/cvpr/GatysEB16}. In our experiments, we also validate the effectiveness of the scheme. This is because person retrieval is a special fine-grained image recognition task, and the network should focus on fine-grained person information. Fusing the features of three levels can bring more detailed features into consideration and thus improve the Re-ID performance. Another appealing property of this network structure is that it can propagate the loss to lower layers in a more efficient way, as lower layers may experience vanishing gradient issues.

\subsection{Ranking Loss}\label{sec:rank_loss}

Currently, for person retrieval, most existing deep-learning-based methods use classification loss or verification loss (e.g., triplet loss and contrastive loss) to train networks. However, person retrieval is in fact a ranking task. Hence, an anchor sample (a query image) needs to interact with multiple positive and negative samples in the retrieval process. We expect that samples having the same identity with the anchor sample should be ranked at top positions. Inspired by the N-pair loss in \cite{DBLP:conf/nips/Sohn16}, a novel ranking loss is developed for person retrieval. Define a batch for training a deep network, as $\mathcal{B}=\{I_{1}, ..., I_{|\mathcal{B}|}\}$. For any $I_{k}, k=1, 2, ...,|\mathcal{B}|$, let $\mathcal{B}^{+}$ and $\mathcal{B}^{-}$ denote positive and negative image sets in $\mathcal{B}$, respectively. $\{x_{1}, ..., x_{|\mathcal{B}|}\}=\{f(I_{1}), ..., f(I_{|\mathcal{B}|})\}$, $f(x_{k})$ is the feature vector normalized by $L2$-norm. The N-pair loss for an anchor $x_{k}$ is defined as
\begin{equation}\label{eq03}
\begin{aligned}
  &\mathcal{L}_{\textrm{N-pair}}(x_{k})=\\
  &\textrm{log}\Big(1+\sum_{j:x_{j}^{-}\in \mathcal{B}^{-}}\textrm{exp}\big(\mathcal{S}(x_{k},x_{j}^{-})-\mathcal{S}(x_{k},x_{k}^{+})\big)\Big),
  \end{aligned}
\end{equation}
where $\mathcal{S}(x_{i},x_{j})=x_{i}^{T}x_{j}$ denotes the similarity between two images.
$x_{k}^{+}$ and $x_{j}^{-}$ are positive and negative samples with $x_{k}$, respectively. Different from existing contrastive loss and triplet loss, the N-pair loss interacts with multiple negative samples simultaneously, while it is only influenced by one positive sample at each iteration.

Considering the ranking process of person retrieval needs to consider not only multiple negative samples, but also multiple positive samples, we formulate a new ranking loss as
\begin{equation}\label{eq04}
\begin{aligned}
  &\mathcal{L}_{\textrm{Rank}}(x_{k})=\\
  &\textrm{log}\Big(1+\sum_{i:x_{i}\in \mathcal{B}^{+} }\sum_{j:x_{j}\in \mathcal{B}^{-}}\textrm{exp}\big(\mathcal{S}(x_{k},x_{j})-\mathcal{S}(x_{k},x_{i})\big)\Big).
  \end{aligned}
\end{equation}
However, Eq.~(\ref{eq04}) needs to calculate $|\mathcal{B}^{+}|\times|\mathcal{B}^{-}|$ pairs of samples in a training batch. To reduce the computational cost, the most dissimilar positive sample  of the query sample is chosen as a reference sample. Meanwhile, to prevent overfitting (\emph{i.e.}, too much attention is paid on predicted correct samples in the ranking), we only select some negative samples which have large similarity with the anchor. Moreover, considering that the similarity of the same identity should be large, we force the similarity between all positive samples and the anchor to be close to one (the largest possible similarity value, as features have been normalized by $L2$-norm). Thus, by rewriting Eq.~(\ref{eq04}), we get the final ranking loss for the proposed neural network as
\begin{equation}\label{eq06}
\begin{aligned}
  &\mathcal{L}_{\textrm{Rank}}(x_{k})=\\
  &\textrm{log}\Big(1+\sum_{j:x_{j}\in \mathcal{B}^{-}}[\textrm{exp}\big(\mathcal{S}(x_{k},x_{j})-\min_{i:x_{i}\in \mathcal{B}^{+}}\mathcal{S}(x_{k},x_{i})+\alpha\big)]_{1+}\Big) \\
  &+\frac{\lambda}{2|\mathcal{B^{+}}|}\sum_{i:x_{i}\in \mathcal{B}^{+} }\big( \mathcal{S}(x_{k},x_{i})-1\big)^{2},
  \end{aligned}
\end{equation}
where $[t]_{1+}$ denotes that if $t>1$, it is equal to $t$, otherwise, 0. $\alpha$ is the margin. In Eq.~(\ref{eq06}), the first term guarantees the negative samples and the most dissimilar positive samples have a margin. The objective of the second term is to make all positive samples similar with the query image. $\lambda$ is a parameter to balance the two terms.

\begin{figure*}%[!ht]
\centering
\subfigure[Triplet loss]{
\centering
\raisebox{15pt}{
\includegraphics[width=3cm]{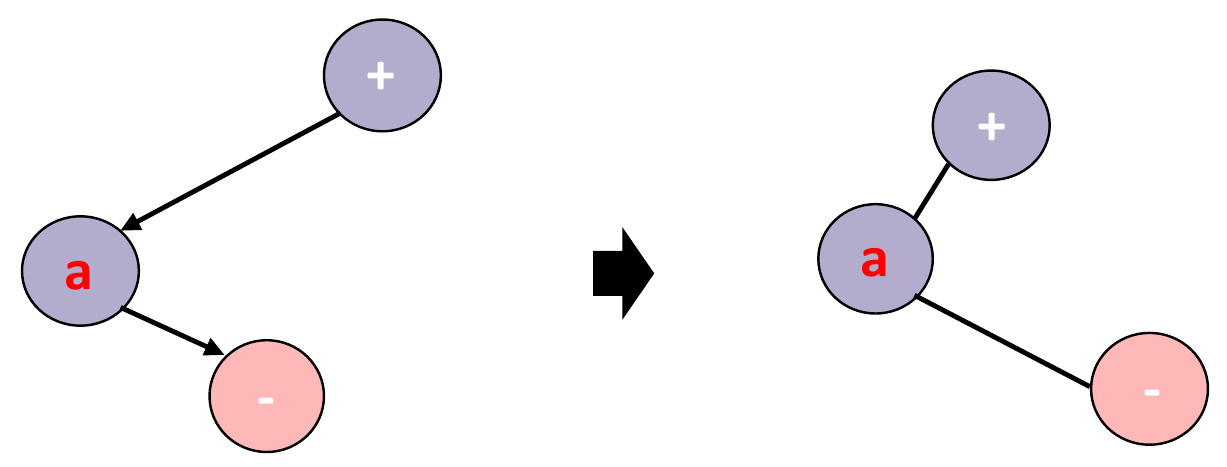}}
}
\subfigure[N-pair loss]{
\centering
\includegraphics[width=4.5cm]{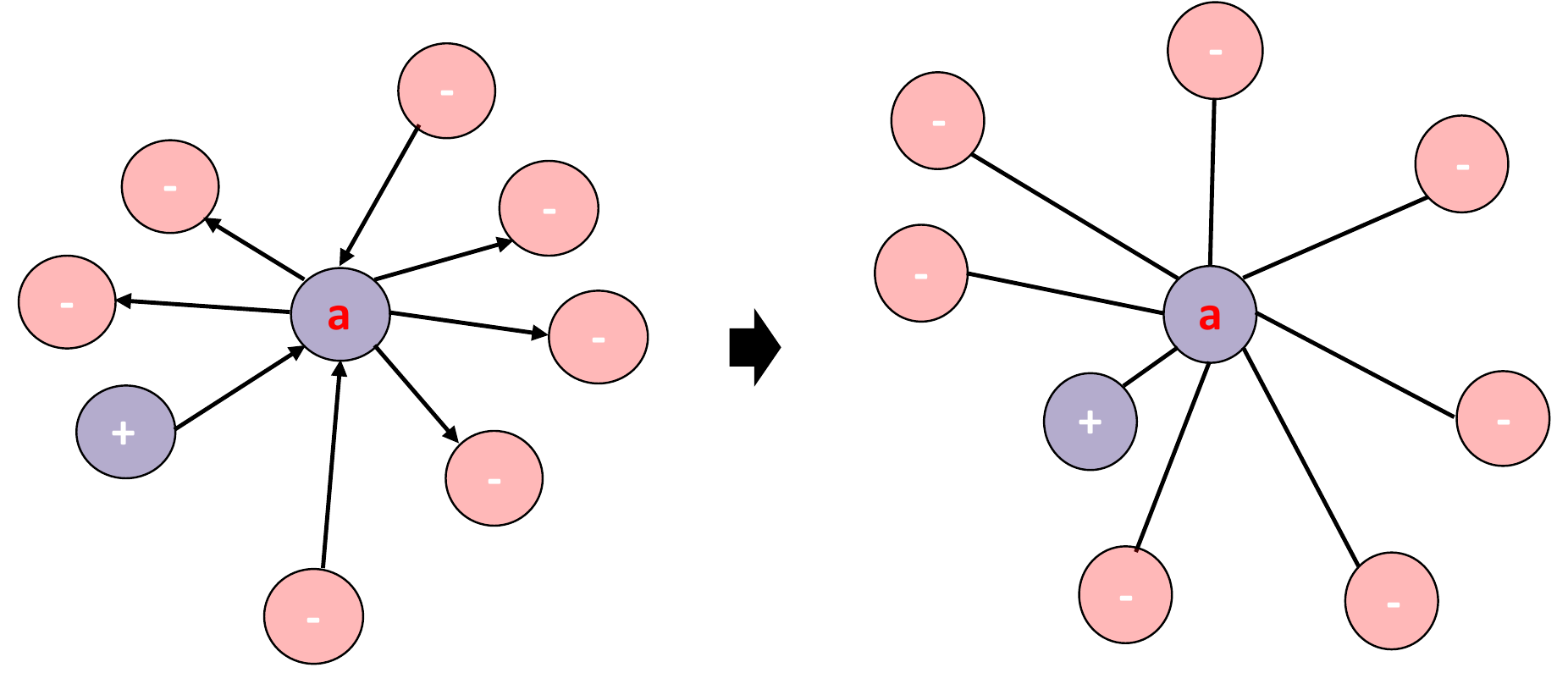}
}
\subfigure[The proposed ranking loss]{
\centering
\includegraphics[width=4.5cm]{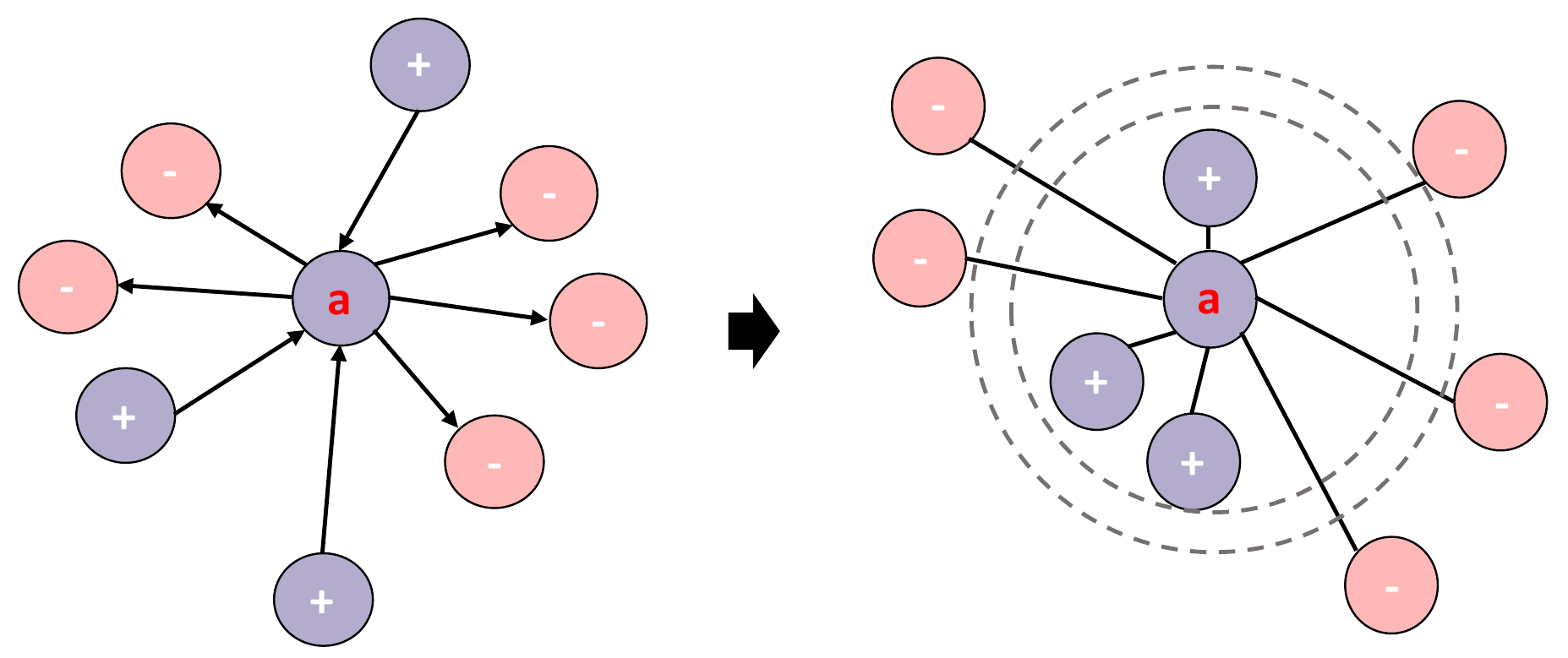}
}
\caption{Illustration of triplet loss, N-pair loss and the proposed ranking loss. For triplet loss, the anchor only interacts with one positive sample and one negative sample. Different from triplet loss, N-pair loss considers that an anchor interacts with one positive sample and multiple negative samples. Moreover, an anchor interacts with multiple positive and negative samples via the proposed ranking loss at each iteration.}
\label{fig4}
\vspace*{-15pt}
\end{figure*}

Compared with the conventional verification loss, as shown in Fig.~\ref{fig4}, the advantage of the proposed ranking loss is that it simultaneously considers multiple positive and negative samples for an anchor at each iteration. Thus, it can make the same ID images from different views closer to query images, and different ID images become farther.

\section{Experiments}
\subsection{Datasets and Evaluation Protocol}\label{exp1}
In this paper, we utilize multiple datasets including small-scale and large-scale datasets to validate the effectiveness of the proposed method.
These small-scale datasets with few persons are collected by few cameras, such as VIPeR, 3DPeS, iLIDS, PRID, Shinpuhkan and CUHK01, as described in~\cite{DBLP:conf/cvpr/XiaoLOW16}.

In recent years, with the extensive application of deep learning in Re-ID, several large-scale datasets have been published. CUHK03 \cite{DBLP:conf/cvpr/LiZXW14} consists of five different camera views and more than $14,000$ images of $1,467$ person. Market1501~\cite{DBLP:conf/iccv/ZhengSTWWT15} contains $32,668$ images of $1,501$ persons. As defined by~\cite{DBLP:conf/iccv/ZhengSTWWT15}, the dataset is split into training$/$testing sets of $12,936/19,732$ images. DukeMTMC-reID~\cite{ristani2016performance} has $16,522$ training images with $1,404$ identities. $2,228$ queries and $17,661$ gallery images are used for evaluation.

Rank-1 accuracy of CMC and mAP are adopted for performance evaluation on Market1501 and DukeMTMC-reID~\cite{DBLP:conf/iccv/ZhengSTWWT15}. We only report Rank-1 accuracy on CUHK03 and all small-scale datasets. Following most of the related work~\cite{DBLP:conf/cvpr/LiC0H17}\cite{DBLP:conf/cvpr/BaiBT17}, all experiments on Market1501 are performed under single query and multiple query settings.

\subsection{Implementation Details}\label{exp2}
Our experiments are done by Caffe with 1080ti (11GB).  All small-scale datasets and CUHK03 are combined together to train a model. The scheme to divide the datasets into the training and testing sets is the same with \cite{DBLP:conf/cvpr/XiaoLOW16}. Since many identities have only two images on these datasets, such as VIPeR, softmax loss is employed for training the proposed deep model. For Market1501 and DukeMTMC-reID, based on the pre-trained model on small-scale datasets, two networks are trained separately with the proposed ranking loss.
For the small-scale datasets, we set the learning rate and iteration to $0.1$ and $55,000$, respectively. In addition, we train the MaskReID with the proposed ranking loss by $20,000$ iterations on Market1501 and DukeMTMC-reID.

 In the training stage, to construct the ranking task, we randomly select one person, then $P$ images with the same identity are chosen randomly. Afterwards, $N$ negative samples are sampled from $N$ person randomly (i.e., one person has only one image). Therefore, in each batch, we can have more negative samples than conventional sampling schemes, which is more reasonable for the retrieval task. In the experiments, we set $P/N$ to $10/54$. Especially, if the chosen person has only $k < 10$ images, $P/N$ is set to $k/(64-k)$.

%In some existing work \cite{DBLP:journals/corr/HermansBL17}, selecting images to form batches is done by randomly sampling $P$ classes (\emph{i.e.}, person identities), and then randomly sampling $K$ images from each class (person), resulting in a batch of $P\times K$ images. In this paper, we randomly select one person, then $P$ images with the same identity are chosen randomly. Afterwards, $N$ negative samples are sampled from $N$ person randomly (\emph{i.e.}, one person has only one image). Therefore, in one batch, we can have more negative samples than conventional sampling schemes, which is more reasonable for a ranking task. In the experiment, we set $P/N$ to 10/54.

\subsection{Comparison with Related Methods}\label{exp3}

\renewcommand{\cmidrulesep}{0mm} %¶¨ÒåÁ½ÌõÏàÁÚ\cmidruleÖ®¼äµÄ¼ä¸ô
\setlength{\aboverulesep}{0mm} %ÔÚÏßÌõ[²»°üÀ¨\toprule]ÉÏÃæÔö¼ÓÒ»¶Î´¹Ö±¾àÀë£¬´Ë´¦Îª0mm
\setlength{\belowrulesep}{0mm} %ÔÚÏßÌõ[²»°üÀ¨\bottomrule]ÏÂÃæÔö¼ÓÒ»Ìõ´¹Ö±¾àÀë£¬´Ë´¦Îª0mm
\setlength{\abovetopsep}{0cm}  %ÔÚÏßÌõ\topruleÉÏÃæ£¬¼´±í¸ñÓëÉÏÃæµÄÎÄ×ÖÖ®¼äµÄ¾àÀë¡£
\setlength{\belowbottomsep}{0cm}%ÔÚÏßÌõ\bottomrule ÏÂÃæ£¬¼´±í¸ñÓëÏÂÃæµÄÎÄ×ÖÖ®¼äµÄ¾àÀë¡£
%\vspace*{-15pt}%
% Table generated by Excel2LaTeX from sheet 'ÎÄÕÂ±í¸ñ'
\begin{table}%[htbp]
  \centering
  \caption{Comparison with the state-of-the-art methods on some benchmark datasets. Note that $1^{st}/2^{nd}$ best in red/blue.}
    \begin{adjustbox}{width=8.5cm}
    \begin{tabular}{|c|c|c|c|c|c|c|c|}
    \toprule
    \textbf{Method} & \textbf{VIPeR} & \textbf{PRID} & \textbf{3DPeS} & \textbf{ iLIDS} & \textbf{CUHK01} & \textbf{CUHK03} \\
    \midrule
    MGCNN~\cite{DBLP:conf/eccv/VariorHW16} & 37.80  & --    & --    & --    & --    & 68.10   \\
    SLSTM~\cite{DBLP:conf/eccv/VariorSLXW16} & 42.40  & --    & --    & --    & --    & 57.30   \\
    DGD~\cite{DBLP:conf/cvpr/XiaoLOW16}   & 38.60  & 64.00  & 56.00  & 64.60  & 66.60  & 75.30  \\
    Spindle~\cite{DBLP:conf/cvpr/ZhaoTSSYYWT17} & \textcolor[rgb]{1.000, 0.000, 0.000}{ \textbf{53.80} } & \textcolor[rgb]{0.000, 0.000, 1.000}{ \textbf{67.00} } & 62.10  & \textcolor[rgb]{0.000, 0.000, 1.000}{\textbf{66.30}}  & 79.90  & \textcolor[rgb]{0.000, 0.000, 1.000}{\textbf{88.50}}   \\
    \midrule
    TCP~\cite{DBLP:conf/cvpr/ChengGZWZ16}   & 47.80  & --    & --    & --    & 53.70  & --     \\
    P2S~\cite{DBLP:conf/cvpr/ZhouWWGZ17}   & --    & --    & \textcolor[rgb]{1.000, 0.000, 0.000}{ \textbf{71.16} } & --    & 77.34  &   --     \\
    Quadruplet~\cite{DBLP:conf/cvpr/ChenCZH17} & 49.05  & --    & --    & --    & \textcolor[rgb]{0.000, 0.000, 1.000}{\textbf{81.00}}  & 75.53  \\
    \midrule
    MaskReID (Ours) & 45.57  & \textcolor[rgb]{1.000, 0.000, 0.000}{ \textbf{70.00} } & \textcolor[rgb]{0.000, 0.000, 1.000}{ \textbf{68.60} } & \textcolor[rgb]{1.000, 0.000, 0.000}{ \textbf{70.43} } & \textcolor[rgb]{1.000, 0.000, 0.000}{ \textbf{84.05} } & \textcolor[rgb]{1.000, 0.000, 0.000}{ \textbf{92.25} } \\
    \bottomrule
    \end{tabular}%
    \end{adjustbox}
  \label{tab1}%
  \vspace*{-15pt}
\end{table}%
\textbf{Small-Scale Datasets.}
For the small-scale datasets, all datasets are combined to train a model. The settings are the same with \cite{DBLP:conf/cvpr/XiaoLOW16}. Since several persons only have two images, softmax loss is used for training our network (MaskReID). The proposed method is compared with different deep learning methods~\cite{DBLP:conf/eccv/VariorHW16}\cite{DBLP:conf/eccv/VariorSLXW16}\cite{DBLP:conf/cvpr/XiaoLOW16}\cite{DBLP:conf/cvpr/ZhaoTSSYYWT17}, and different loss function based methods~\cite{DBLP:conf/cvpr/ChengGZWZ16}\cite{DBLP:conf/cvpr/ZhouWWGZ17}\cite{DBLP:conf/cvpr/ChenCZH17}.
Experimental results are reported in Table~\ref{tab1}. 
As seen, some observations can be made as follows: i) Performance varies on different datasets. In particular, the Rank-1 is up to $92.25\%$ on the CUHK03. This may be because CUHK03 has more training images than other datasets.  However, the relatively poor performance on VIPeR is partly because the images of VIPeR are of low resolution which makes the segmentation results poor; ii) The proposed method can achieve competitive results when compared with the state-of-the-art methods on several benchmark datasets. This confirms the effectiveness of our proposed MaskReID.

\begin{table}%[!htbp]
  \centering
  \caption{Comparison with the state-of-the-art methods on Market1501. Note that $1^{st}/2^{nd}$ best in red/blue.}
  \begin{adjustbox}{width=8cm}
    \begin{tabular}{|c|cc|cc|c|}
    \toprule
          & \multicolumn{2}{c|}{~~Single query~~} & \multicolumn{2}{c|}{~~Multiple query~~}  \\
    \midrule
    Method & Rank-1 & mAP   & Rank-1 & mAP   \\
    \midrule
    DLPAR~\cite{DBLP:conf/iccv/ZhaoLZW17} & 81.00    & 63.40  & --    & --     \\
    Spindle~\cite{DBLP:conf/cvpr/ZhaoTSSYYWT17} & 76.90  & --    & --    & --     \\
    MSCAN~\cite{DBLP:conf/cvpr/LiC0H17}& 80.31 & 57.53 & 86.79 & 66.70   \\
    S2S~\cite{zhou2018large}   & 65.32 & 39.83 & 80.49 & 52.69  \\
    P2S~\cite{DBLP:conf/cvpr/ZhouWWGZ17}   & 70.72 & 44.27 & 85.78 & 55.73  \\
    SSM~\cite{DBLP:conf/cvpr/BaiBT17}   & 82.21 & 68.80  & 88.18 & 76.18  \\
    JLML~\cite{DBLP:conf/ijcai/LiZG17}  & 83.90  & 64.40  & 89.70  & 74.50   \\
    \midrule
    MaskReID (Ours) &   \textcolor[rgb]{0.000, 0.000, 1.000}{\textbf{90.44}}    &   \textcolor[rgb]{0.000, 0.000, 1.000}{\textbf{75.36}}    &   \textcolor[rgb]{0.000, 0.000, 1.000}{\textbf{93.35}}    &  \textcolor[rgb]{0.000, 0.000, 1.000}{\textbf{82.37}}      \\
	    MaskReID{\tiny re-ranking} (Ours)&   \textcolor[rgb]{1.000, 0.000, 0.000}{\textbf{92.46}}    &   \textcolor[rgb]{1.000, 0.000, 0.000}{\textbf{88.13}}    &  \textcolor[rgb]{1.000, 0.000, 0.000}{\textbf{94.77}}     &   \textcolor[rgb]{1.000, 0.000, 0.000}{\textbf{92.11}}     \\
    \bottomrule
    \end{tabular}%
    \end{adjustbox}
  \label{tab2}%
  \vspace*{-10pt}
\end{table}%

% Table generated by Excel2LaTeX from sheet 'ÎÄÕÂ±í¸ñ'
%\vspace*{-15pt}%
\begin{table}%[!htbp]
  \centering
  \caption{Comparison with the state-of-the-art methods on DukeMTMC-reID. Note that $1^{st}/2^{nd}$ best in red/blue.}
    \begin{tabular}{|c|cc|c|}
    \toprule
    Method & ~~Rank-1~~ & ~~mAP~~  \\
    \midrule
    LSRO~\cite{DBLP:conf/iccv/ZhengZY17}   & 67.68  & 47.13  \\
    SVDNet~\cite{DBLP:conf/iccv/SunZDW17} & 76.70  & 56.80 \\
    OIM~\cite{DBLP:conf/cvpr/XiaoLWLW17}   & 68.10  & --     \\
    ACRN~\cite{DBLP:conf/cvpr/SchumannS17}  & 72.58  & 51.96   \\
    \midrule
    MaskReID (Ours) &  \textcolor[rgb]{0.000, 0.000, 1.000}{\textbf{78.86}}     &   \textcolor[rgb]{0.000, 0.000, 1.000}{\textbf{61.89}}    \\
    MaskReID{\tiny re-ranking} (Ours)&   \textcolor[rgb]{1.000, 0.000, 0.000}{\textbf{84.07}}    &   \textcolor[rgb]{1.000, 0.000, 0.000}{\textbf{79.73}}     \\
    \bottomrule
    \end{tabular}%
  \label{tab3}%
  \vspace*{-10pt}
\end{table}%

\textbf{Large-Scale Datasets.}
The proposed method is also validated on Market1501~\cite{DBLP:conf/iccv/ZhengSTWWT15} and DukeMTMC-reID~\cite{ristani2016performance}. In this experiment, the proposed method is compared with a set of the state-of-the-art methods, i.e., DLPAR~\cite{DBLP:conf/iccv/ZhaoLZW17}, Spindle~\cite{DBLP:conf/cvpr/ZhaoTSSYYWT17}, MSCAN~\cite{DBLP:conf/cvpr/LiC0H17}, S2S~\cite{zhou2018large}, P2S~\cite{DBLP:conf/cvpr/ZhouWWGZ17}, SSM~\cite{DBLP:conf/cvpr/BaiBT17} and JLML~\cite{DBLP:conf/ijcai/LiZG17}.
Experimental results are given in Table~\ref{tab2} and~\ref{tab3}.
From the two tables, we can observe that: i) Comparison with several part-based methods, such as MSCAN~\cite{DBLP:conf/cvpr/LiC0H17} and JLML~\cite{DBLP:conf/ijcai/LiZG17}, the proposed method has a competitive performance. Although MaskReID is a global method which extracts features from the global image, it attends more on the foreground region and fuses features of  different layers to obtain the low-level detailed information and high-level semantic information from person images. Moreover, the proposed method outperforms DLPAR~\cite{DBLP:conf/iccv/ZhaoLZW17} that used attention mechanism and Spindle~\cite{DBLP:conf/cvpr/ZhaoTSSYYWT17} that employed human pose information; ii) The proposed method consistently outperforms the state-of-the-art methods on large-scale datasets. Especially, on Market1501, Rank-1$/$mAP of the multiple query is up to $93.35/82.37\%$. As demonstrated, it can effectively learn a more discriminative feature representation for person retrieval;
iii) Employing the re-ranking algorithm~\cite{DBLP:conf/cvpr/ZhongZCL17} can further improve the performance of our proposed method. Re-ranking algorithm~\cite{DBLP:conf/cvpr/ZhongZCL17} is commonly used for Re-ID, which can further enhance the performance of person retrieval via exploring the relationship of each sample in the ranking list. As can be seen in both Table~\ref{tab2} and~\ref{tab3}, performance can be further improved. On Market1501, Rank-1$/$mAP of the single query is now up to $92.46/88.13\%$.

% Table generated by Excel2LaTeX from sheet 'ÎÄÕÂ±í¸ñ'
%\vspace*{-15pt}%
\begin{table}%[htbp]
  \centering
  \caption{Performance of the proposed method when employing different network components on many benchmark datasets. Note that $1^{st}/2^{nd}$ best in red/blue.}
  \begin{adjustbox}{width=8.5cm}
    \begin{tabular}{|c|c|c|c|c|c|c|}
    \toprule
    \textbf{Method} & \textbf{VIPeR} & \textbf{PRID} & \textbf{3DPeS} & \textbf{ iLIDS} & \textbf{CUHK01} & \textbf{CUHK03} \\
    \midrule
    DGD   & 38.60  & 64.00  & 56.00  & 64.60  & 66.60  & 75.30  \\
    MaskReID-F & 39.24  & 64.00  & 64.88  & 66.09  & 77.16  & 87.47  \\
    MaskReID-M & \textcolor[rgb]{ 0,  0,  1}{\textbf{44.62 }} & \textcolor[rgb]{ 0,  0,  1}{\textbf{65.00 }} & \textcolor[rgb]{ 0,  0,  1}{\textbf{66.12 }} & \textcolor[rgb]{ 0,  0,  1}{\textbf{69.57 }} & \textcolor[rgb]{ 1,  0,  0}{\textbf{84.26 }} & \textcolor[rgb]{ 0,  0,  1}{\textbf{88.75 }} \\
    \midrule
    MaskReID (Ours) & \textcolor[rgb]{ 1,  0,  0}{\textbf{45.57 }} & \textcolor[rgb]{ 1,  0,  0}{\textbf{70.00 }} & \textcolor[rgb]{ 1,  0,  0}{\textbf{68.60 }} & \textcolor[rgb]{ 1,  0,  0}{\textbf{70.43 }} & \textcolor[rgb]{ 0,  0,  1}{\textbf{84.05 }} & \textcolor[rgb]{ 1,  0,  0}{\textbf{92.25 }} \\
    \bottomrule
    \end{tabular}%
    \end{adjustbox}
  \label{tab5}%
  \vspace*{-10pt}
\end{table}%
\subsection{Ablation Studies}
\textbf{Effectiveness of Different Network Components.}
In this section, we validate the effectiveness of different network components. Table~\ref{tab5} shows the experimental results on multiple datasets. DGD is the original network framework.  MaskReID is built on DGD with both masked image as input and the multi-level feature fusion. Compared with MaskReID, MaskReID-F and MaskReID-M denote the removal of the input masked images and multi-level feature fusion structure, respectively. By comparing with the reports on multiple datasets in Table~\ref{tab5}, some observations can be made as follows: i) The fusion of multi-layer features is effective for the person Re-ID, i.e., MaskReID-F outperforms the original DGD. This demonstrates the effectiveness of the multi-layer feature fusion in the proposed framework; ii) Using masked images can improve the performance of person Re-ID, i.e., MaskReID-M outperforms the original DGD. Employing masked images can reduce the impact of background clutter, and only focus on the person region; iii) MaskReID, including masked input and stacking the multi-layer features, further improves the performance of the person Re-ID. In summary, the experimental results are consistent with the analysis in Section~\ref{sec:framework}.

\begin{table}[htbp]
  \centering
  \caption{Evaluation of different loss functions on Market1501.}
    \begin{tabular}{|c|cc|cc|}
  %  \toprule
%    ~~Loss function~~ & \multicolumn{2}{c|}{~~Single query~~} & \multicolumn{2}{c|}{~~Multiple query~~} \\
%    \midrule
      \toprule
          & \multicolumn{2}{c|}{~~Single query~~} & \multicolumn{2}{c|}{~~Multiple query~~}  \\
    \midrule
    ~~Loss function~~ & Rank-1 & mAP   & Rank-1 & mAP   \\
    \midrule
    Softmax loss & 88.18 & 70.57 & 91.48 & 78.06 \\
    Triplet loss & 88.54  & 71.17 & 92.13 & 78.66 \\
    N-pair loss & 89.52 & 73.09 & 91.66 & 79.17 \\
    \midrule
    Ranking loss (Ours) & \textcolor[rgb]{1,  0,  0}{\textbf{90.44}} & \textcolor[rgb]{ 1,  0,  0}{\textbf{75.36}} & \textcolor[rgb]{ 1,  0,  0}{\textbf{93.35}} & \textcolor[rgb]{ 1,  0,  0}{\textbf{82.37}} \\
    \bottomrule
    \end{tabular}%
  \label{tab6}%
  \vspace*{-10pt}
\end{table}%
\begin{table*}[htbp]
  \centering
  \caption{Performance of the proposed method when setting different $\alpha$ and $\lambda$ on Market1501. Note that red/green denotes the best/worst result.}
  %\begin{adjustbox}{width=8.5cm}
    \begin{tabular}{|c|cc|cc|cc|cc|cc|}
    \toprule
     & \multicolumn{2}{c|}{$\alpha$=0.1} & \multicolumn{2}{c|}{$\alpha$=0.15} & \multicolumn{2}{c|}{$\alpha$=0.2} & \multicolumn{2}{c|}{$\alpha$=0.5} & \multicolumn{2}{c|}{$\alpha$=1.0} \\
    \midrule
    $\lambda$ & Rank-1 & mAP   & Rank-1 & mAP   & Rank-1 & mAP   & Rank-1 & mAP   & Rank-1 & mAP \\
    \midrule
    0     & \textcolor[rgb]{ 0,  .69,  .314}{\textbf{89.79 }} & \textcolor[rgb]{ 0,  .69,  .314}{\textbf{72.78 }} & \textcolor[rgb]{ 0,  .69,  .314}{\textbf{89.76 }} & \textcolor[rgb]{ 0,  .69,  .314}{\textbf{72.71 }} & 90.11  & \textcolor[rgb]{ 0,  .69,  .314}{\textbf{72.99 }} & 89.46  & 74.19  & \textcolor[rgb]{ 1,  0,  0}{\textbf{79.87 }} & \textcolor[rgb]{ 1,  0,  0}{\textbf{58.87 }} \\
    1     & 89.85  & 74.97  & 89.93  & 75.22  & \textcolor[rgb]{ 1,  0,  0}{\textbf{90.44 }} & \textcolor[rgb]{ 1,  0,  0}{\textbf{75.36 }} & \textcolor[rgb]{ 1,  0,  0}{\textbf{89.55 }} & \textcolor[rgb]{ 1,  0,  0}{\textbf{74.60 }} & 79.72  & 57.81  \\
    2     & \textcolor[rgb]{ 1,  0,  0}{\textbf{90.05 }} & 75.12  & \textcolor[rgb]{ 1,  0,  0}{\textbf{90.02 }} & \textcolor[rgb]{ 1,  0,  0}{\textbf{75.30 }} & 90.17  & 75.27  & 89.04  & 73.74  & 79.66  & 57.33  \\
    5     & 89.90  & \textcolor[rgb]{ 1,  0,  0}{\textbf{75.23 }} & 89.88  & 75.04  & 89.31  & 74.54  & 87.44  & 71.75  & 78.95  & 58.03  \\
    10    & 89.88  & 74.90  & 89.22  & 74.42  & \textcolor[rgb]{ 0,  .69,  .314}{\textbf{88.36 }} & 73.36  & \textcolor[rgb]{ 0,  .69,  .314}{\textbf{86.02 }} & \textcolor[rgb]{ 0,  .69,  .314}{\textbf{69.50 }} & \textcolor[rgb]{ 0,  .69,  .314}{\textbf{77.73 }} & \textcolor[rgb]{ 0,  .69,  .314}{\textbf{57.30 }} \\
    \bottomrule
    \end{tabular}%
   %\end{adjustbox}
  \label{tab4}%
  \vspace*{-10pt}
\end{table*}%

\textbf{Effectiveness of the Ranking Loss.}
In this part, we utilize different loss functions, including softmax loss, triplet loss, N-pair loss and the proposed ranking loss, to train the proposed network on Market1501. Specifically, we train our model with triplet loss by online hard sample mining and utilize the same parameters (e.g., the number of iterations, learning rate, batch size, etc.) for all loss functions.
Table \ref{tab6} reports the experimental results. We can observer that: i) N-pair loss outperforms the triplet loss.  Compared with the conventional triplet loss, N-pair loss interacts with one positive sample and multiple negative samples; ii) The proposed method has better performance than N-pair loss and triplet loss. This validates the analysis in Section~\ref{sec:rank_loss}.

\subsection{Evaluation of Parameters in Ranking Loss}
 In this section, parameters of the proposed ranking loss function are analyzed, i.e., parameters in Eq.~(\ref{eq06}). %Eq.~(\ref{eq06}) has two terms, where the first term is to force a margin between positive and negative samples and the second term is designed to make all the positive samples closer to the query image. $\lambda$ is a parameter to trade off the two terms. $\alpha$ is the margin which controls the distances between positive and negative samples. Several experiments are carried out with different $\lambda$ and $\alpha$. Results are reported in Table~\ref{tab4}. In particular, we want to validate the effectiveness of each term. For $\alpha$, it is set to 0.1, 0.15, 0.2, 0.5 and 1.0. $\lambda$ is set to $0, 1, 2, 5, 10$.
From Table~\ref{tab4}, we can observe that: i) when $\lambda$ is set to 0, i.e., removing the second item in Eq.~(\ref{eq06}), the performance is poor on mAP. This demonstrates that enhancing the similarity between all the positive images and the query image in the optimization process is useful; ii) In addition, when $\lambda$ increases to values larger than 2, the performance decreases slightly. This implies that focusing too much on the second term of Eq.~(\ref{eq06}) could hurt the performance. We need a reasonable $\lambda$ to balance the two terms in Eq.~(\ref{eq06}); iii) The results of setting $\alpha$ to 0.1, 0.15 and 0.2 are similar, with the results of 0.2 slightly better. As a margin parameter, $\alpha$ should not be too large as an over-large $\alpha$ may easily lead to overfitting problems. When $\alpha$ is over-large (say $\geq 0.5$), there will be excessive penalty from the loss function while optimizing the weights of the proposed network. In conclusion, we set $\lambda$/$\alpha$ to $1/0.2$ in all experiments.
\section{Conclusion}
In this paper, a mask based deep ranking neural network is developed to deal with person retrieval. First, to reduce the impact of messy background in different camera views, the masked images together with the original images are used as input. Second, to obtain more discriminative information, we can combine low-, middle- and high-level features to form a merged feature. Third, we put forward a novel ranking loss function to optimize the weights of the network to further alleviate the interference of similar appearance in person retrieval. Results on various datasets, including both small-scale and large-scale datasets, show the effectiveness of the proposed method compared with a set of state-of-the-art methods.%The proposed method outperforms the state-of-the-art methods on many large-scale datasets.

%Since the part model can reduce the interference of variance of human pose in different views, several part models are proposed to deal with the person Re-ID problem. MaskReID is a global based method, \emph{i.e.}, employing the whole image as input, without considering body part information. In the future, we will exploit some strategies to combine the masked image and the body part information to further improve the performance of the person Re-ID task. In addition, to migrate the impact of the poor segmentation, we will also add an image-adaptive network module to control if the segmentation mask of this image shall be considered or neglected in the input.

\bibliographystyle{IEEEbib}
\bibliography{egbib}

\begin{thebibliography}{10}

\bibitem{ristani2016performance}
Ergys Ristani, Francesco Solera, Roger Zou, Rita Cucchiara, and Carlo Tomasi,
\newblock ``Performance measures and a data set for multi-target, multi-camera
  tracking,''
\newblock in {\em ECCV}, 2016.

\bibitem{DBLP:conf/eccv/VariorSLXW16}
Rahul~Rama Varior, Bing Shuai, Jiwen Lu, Dong Xu, and Gang Wang,
\newblock ``A siamese long short-term memory architecture for human
  re-identification,''
\newblock in {\em ECCV}, 2016.

\bibitem{DBLP:conf/cvpr/XiaoLOW16}
Tong Xiao, Hongsheng Li, Wanli Ouyang, and Xiaogang Wang,
\newblock ``Learning deep feature representations with domain guided dropout
  for person re-identification,''
\newblock in {\em CVPR}, 2016.

\bibitem{DBLP:conf/cvpr/ZhaoTSSYYWT17}
Haiyu Zhao, Maoqing Tian, Shuyang Sun, Jing Shao, Junjie Yan, Shuai Yi,
  Xiaogang Wang, and Xiaoou Tang,
\newblock ``Spindle net: Person re-identification with human body region guided
  feature decomposition and fusion,''
\newblock in {\em CVPR}, 2017.

\bibitem{DBLP:conf/cvpr/ChengGZWZ16}
De~Cheng, Yihong Gong, Sanping Zhou, Jinjun Wang, and Nanning Zheng,
\newblock ``Person re-identification by multi-channel parts-based {CNN} with
  improved triplet loss function,''
\newblock in {\em CVPR}, 2016.

\bibitem{DBLP:conf/cvpr/ZhouWWGZ17}
Sanping Zhou, Jinjun Wang, Jiayun Wang, Yihong Gong, and Nanning Zheng,
\newblock ``Point to set similarity based deep feature learning for person
  re-identification,''
\newblock in {\em CVPR}, 2017.

\bibitem{DBLP:conf/cvpr/ChenCZH17}
Weihua Chen, Xiaotang Chen, Jianguo Zhang, and Kaiqi Huang,
\newblock ``Beyond triplet loss: {A} deep quadruplet network for person
  re-identification,''
\newblock in {\em CVPR}, 2017.

\bibitem{DBLP:conf/eccv/VariorHW16}
Rahul~Rama Varior, Mrinal Haloi, and Gang Wang,
\newblock ``Gated siamese convolutional neural network architecture for human
  re-identification,''
\newblock in {\em ECCV}, 2016.

\bibitem{DBLP:conf/cvpr/LongSD15}
Jonathan Long, Evan Shelhamer, and Trevor Darrell,
\newblock ``Fully convolutional networks for semantic segmentation,''
\newblock in {\em CVPR}, 2015.

\bibitem{DBLP:conf/cvpr/GatysEB16}
Leon~A. Gatys, Alexander~S. Ecker, and Matthias Bethge,
\newblock ``Image style transfer using convolutional neural networks,''
\newblock in {\em CVPR}, 2016.

\bibitem{DBLP:conf/nips/Sohn16}
Kihyuk Sohn,
\newblock ``Improved deep metric learning with multi-class n-pair loss
  objective,''
\newblock in {\em NeurIPS}, 2016.

\bibitem{DBLP:conf/cvpr/LiZXW14}
Wei Li, Rui Zhao, Tong Xiao, and Xiaogang Wang,
\newblock ``Deepreid: Deep filter pairing neural network for person
  re-identification,''
\newblock in {\em CVPR}, 2014.

\bibitem{DBLP:conf/iccv/ZhengSTWWT15}
Liang Zheng, Liyue Shen, Lu~Tian, Shengjin Wang, Jingdong Wang, and Qi~Tian,
\newblock ``Scalable person re-identification: {A} benchmark,''
\newblock in {\em ICCV}, 2015.

\bibitem{DBLP:conf/cvpr/LiC0H17}
Dangwei Li, Xiaotang Chen, Zhang Zhang, and Kaiqi Huang,
\newblock ``Learning deep context-aware features over body and latent parts for
  person re-identification,''
\newblock in {\em CVPR}, 2017.

\bibitem{DBLP:conf/cvpr/BaiBT17}
Song Bai, Xiang Bai, and Qi~Tian,
\newblock ``Scalable person re-identification on supervised smoothed
  manifold,''
\newblock in {\em CVPR}, 2017.

\bibitem{DBLP:conf/iccv/ZhaoLZW17}
Liming Zhao, Xi~Li, Yueting Zhuang, and Jingdong Wang,
\newblock ``Deeply-learned part-aligned representations for person
  re-identification,''
\newblock in {\em ICCV}, 2017.

\bibitem{zhou2018large}
Sanping Zhou, Jinjun Wang, Rui Shi, Qiqi Hou, Yihong Gong, and Nanning Zheng,
\newblock ``Large margin learning in set-to-set similarity comparison for
  person reidentification,''
\newblock {\em TMM}, 2018.

\bibitem{DBLP:conf/ijcai/LiZG17}
Wei Li, Xiatian Zhu, and Shaogang Gong,
\newblock ``Person re-identification by deep joint learning of multi-loss
  classification,''
\newblock in {\em IJCAI}, 2017.

\bibitem{DBLP:conf/iccv/ZhengZY17}
Zhedong Zheng, Liang Zheng, and Yi~Yang,
\newblock ``Unlabeled samples generated by {GAN} improve the person
  re-identification baseline in vitro,''
\newblock in {\em ICCV}, 2017.

\bibitem{DBLP:conf/iccv/SunZDW17}
Yifan Sun, Liang Zheng, Weijian Deng, and Shengjin Wang,
\newblock ``Svdnet for pedestrian retrieval,''
\newblock in {\em ICCV}, 2017.

\bibitem{DBLP:conf/cvpr/XiaoLWLW17}
Tong Xiao, Shuang Li, Bochao Wang, Liang Lin, and Xiaogang Wang,
\newblock ``Joint detection and identification feature learning for person
  search,''
\newblock in {\em CVPR}, 2017.

\bibitem{DBLP:conf/cvpr/SchumannS17}
Arne Schumann and Rainer Stiefelhagen,
\newblock ``Person re-identification by deep learning attribute-complementary
  information,''
\newblock in {\em CVPRW}, 2017.

\bibitem{DBLP:conf/cvpr/ZhongZCL17}
Zhun Zhong, Liang Zheng, Donglin Cao, and Shaozi Li,
\newblock ``Re-ranking person re-identification with k-reciprocal encoding,''
\newblock in {\em CVPR}, 2017.

\end{thebibliography}

\end{document}